\documentclass[letterpaper, 10 pt, conference]{ieeeconf}  

\IEEEoverridecommandlockouts                              
\usepackage{times}
\usepackage{multicol}
\usepackage[bookmarks=true]{hyperref}
\usepackage{algorithm}
\usepackage[noend]{algpseudocode}
\usepackage{mathtools}
\usepackage{amsfonts}
\usepackage{amssymb}
\usepackage{bm}
\usepackage{mathrsfs}
\usepackage{amsmath}
\usepackage{cite}

\usepackage[capitalize]{cleveref} 
\usepackage{multirow}
\usepackage{array}
\usepackage{caption}
\usepackage{makecell}
\usepackage{booktabs}
\usepackage{multirow}
\usepackage{hyperref}
\usepackage{xcolor}
\usepackage{fancyhdr}
\usepackage[switch]{lineno}
\usepackage{soul}
\soulregister{\cref}{7}
\soulregister{\ref}{7}
\soulregister{\cite}{7}
\newtheorem{definition}{Definition}
\newtheorem{problem}{Problem}

\newcommand{\trajectory}{T}
\newcommand{\trajectorydata}{\mathcal{D}}
\newcommand{\textualdata}{\mathcal{D}^{\step \state \control}}
\newcommand{\visualdata}{\mathcal{D}^{\image}}

\newcommand{\context}{\textbf{C}}
\newcommand{\contextLLMtasking}{\textbf{C}^{task}}
\newcommand{\contextdatadescription}{\textbf{C}^{\mathcal{D}}}
\newcommand{\contextoneshot}{\textbf{C}^{1S}}
\newcommand{\contextzeroshot}{\textbf{C}^{0S}}

\newcommand{\prompt}{\textbf{P}}

\newcommand{\Stateset}{X}
\newcommand{\Controlset}{U}
\newcommand{\Imageset}{\mathcal{I}}

\newcommand{\statedim}{p}
\newcommand{\controldim}{q}

\newcommand{\step}{k}
\newcommand{\nsteps}{K}
\newcommand{\state}{x}
\newcommand{\control}{u}
\newcommand{\image}{\mu}
\newcommand{\startstep}{\step_{\text{start}}}
\newcommand{\subtaskdescription}{\zeta}
\newcommand{\nsubtasks}{N}
\newcommand{\idxsubtask}{n}

\newcommand{\subtask}{s}
\newcommand{\Subtaskdecomposition}{\mathcal{S}}

\newcommand{\R}{\mathbb{R}}
\newcommand{\N}{\mathbb{N}}

\newcommand{\SIMILARITY}{\texttt{SIMILARITY}}

\newcommand{\temporalscore}{\tau_{k}}
\newcommand{\semanticscore}{\tau_{\zeta}}
\newcommand{\intervalweight}{w}

\newcommand{\intersectingsubtaskidx}{i}

\newcommand\nohat[1]{#1\vphantom{\hat{#1}}}

\newcommand{\hlmath}[1]{#1}
\newcommand{\hlyellow}[1]{#1}

\newcommand{\hlgreen}[1]{#1}

\title{\LARGE \bf
Temporal and Semantic Evaluation Metrics for \\ Foundation Models in Post-Hoc Analysis of Robotic Sub-tasks
}
\author{Jonathan Salfity$^{1}$, Selma Wanna$^{1}$, Minkyu Choi$^{2}$, and Mitch Pryor$^{1}$
\thanks{$^{1}$Nuclear and Applied Robotics Group, Department of Mechanical Engineering and $^{2}$SWARM Lab, Department of Electrical and Computer Engineering,
        The University of Texas at Austin,
        {\tt\small \{j.salfity, slwanna, minkyu.choi, mpryor\}@utexas.edu}}%
}

\begin{document}
\maketitle

\thispagestyle{empty}
\thispagestyle{fancy}
\renewcommand{\headrulewidth}{0pt}
\renewcommand{\footrulewidth}{0pt}
\fancyfoot[C]{\footnotesize This work has been submitted to the IEEE for possible publication. Copyright may be transferred without notice, after which this version may no longer be accessible.}

\pagestyle{empty}

\begin{abstract}
Recent works in Task and Motion Planning (TAMP) show that training control policies on language-supervised robot trajectories with quality labeled data markedly improves agent task success rates. 
However, the scarcity of such data presents a significant hurdle to extending these methods to general use cases. 
To address this concern, we present an automated framework to decompose trajectory data into temporally bounded and natural language-based descriptive sub-tasks by leveraging recent prompting strategies for Foundation Models (FMs) including both Large Language Models (LLMs) and Vision Language Models (VLMs).
Our framework provides both time-based and language-based descriptions for lower-level sub-tasks that comprise full trajectories.
To rigorously evaluate the quality of our automatic labeling framework, we contribute an algorithm $\SIMILARITY$ to produce two novel metrics: temporal similarity and semantic similarity.
The metrics measure the temporal alignment and semantic fidelity of language descriptions between two sub-task decompositions, namely an FM sub-task decomposition prediction and a ground-truth sub-task decomposition.
We present scores for temporal similarity and semantic similarity above 90\%, compared to $\approx$60\% for human annotations, for multiple, simulated robotic environments, demonstrating the effectiveness of our proposed framework. 
Our results enable building diverse, large-scale, language-supervised datasets for improved robotic TAMP.
\end{abstract}

\section{Introduction}
\label{sec:Introduction}
Training robotic agents on language-supervised robot trajectory data generally improves their success rates for Task and Motion Planning (TAMP) applications \cite{pmlr-v164-jang22a,brohan2023rt1}.
However, the scarcity of large-scale, language-annotated robot trajectory data, particularly at the granular sub-task level, hinders mass adoption of these methods. 
The lack of clean, diverse, large-scale and accessible datasets also hampers the creation of Foundation Models (FMs) \cite{bommasani2022opportunities} tailored to robotic use cases \cite{brohan2023rt1, open_x_embodiment_rt_x_2023}. 
This prevents robotic researchers from reaching an ``ImageNet Moment'' \cite{imagenet}: a moment that revolutionized computer vision by creating a vast and diverse labeled dataset.

\begin{figure}[ht!]
    \includegraphics[width=0.48\textwidth]{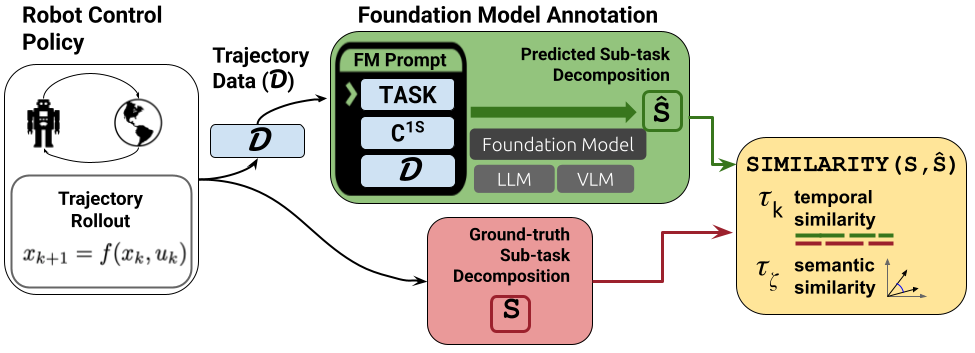}
    \caption{Our approach evaluates a Foundation Model's (FM) ability to temporally and semantically decompose a robot trajectory into a sub-task decomposition. 
    We compare an FM sub-task decomposition, $\hat \Subtaskdecomposition$, with a ground-truth sub-task decomposition, $\Subtaskdecomposition$, through our core contributions of temporal and semantic alignment metrics.
    The image above shows how robot trajectory data, $\trajectorydata$, is post-hoc processed by an FM to compute a predicted sub-task decomposition, $\hat \Subtaskdecomposition$, and quantitatively compared to a ground-truth sub-task decomposition, $\Subtaskdecomposition$.}
  \label{fig:main_flow_diagran}
\end{figure}

Recent works address data scarcity by manually collecting, cleaning, and curating new or existing datasets into large, open-source repositories \cite{pmlr-v164-jang22a, brohan2023rt1, open_x_embodiment_rt_x_2023}. 
However, efforts to scale language annotations, despite their well-known benefits, are often neglected. 
\hlyellow{Prior works that offer automated language annotation methods predominantly generate high-level language instructions for full trajectories \cite{xiao2022robotic} or rely on computationally expensive validation methods \cite{ha2023scaling}.
By contrast, we develop rich, nuanced language descriptions at a granular sub-task level towards improving the robotics and Embodied-AI agents \cite{embodiedai} capabilities in understanding a wider range of TAMP instructions via refined language supervision.}

This paper proposes a \hlgreen{sub-task annotation} framework, see \cref{fig:main_flow_diagran}, which leverages FMs including state-of-the-art Visual Language Models (VLMs) \cite{openai2023gpt4vision, geminiteam2023gemini, lin2023videollava} and Large Language Models (LLMs) to automatically generate natural language descriptions and annotations for robot trajectories at sub-task level.
Compared to traditional manual annotation via platforms like MTurk \cite{mturk}, our approach offers significant advantages: scalability and cost-effectiveness. 
For example, a human annotator earning a minimum wage of \$14.41 per hour \cite{acmDataDrivenAnalysis}, would need to complete each annotation in 15-30 seconds to match the cost-effectiveness of our GPT-4V method; in our experiments, we found \hlgreen{human} annotation took between 2-3 minutes.


\hlgreen{To evaluate the quality of each annotation produced by our framework}, we define two metrics: temporal similarity, $\temporalscore$, and semantic similarity, $\semanticscore$.
Both metrics \hlgreen{are designed to} quantify alignment between a ground-truth annotation and FM-generated annotation, \hlgreen{allowing researchers to assess the quality of the proposed framework on their own robotics' trajectories.}
This methodology only requires a test set (a small sample of labeled data) to evaluate the its performance instead of collecting a dataset for fine-tuning.
Our experimental results, accumulated over 200 unique robot trajectories with 2400 variations of FM calls, demonstrate the proposed framework produces valid annotations with a $\temporalscore$ and $\semanticscore$ above 90\% for contextually informative prompts compared to human-based annotations below 60\%, indicating its effectiveness in adhering to the semantic and temporal realities of the ground-truth.

\hlyellow{
Our core contributions are as follows.}
\begin{itemize}
    \item \hlyellow{A framework to automate trajectory annotations using an FM to ingest raw, multi-modal trajectory data and predict a sub-task decomposition, $\hat \Subtaskdecomposition$.}
    \item \hlyellow{Similarity metrics, $\temporalscore$ and $\semanticscore$, to quantitatively evaluate the temporal and semantic alignment between two sub-task decompositions: an actual $\Subtaskdecomposition$ and prediction $\hat \Subtaskdecomposition$.}
    \item \hlyellow{Results of the metrics across concrete examples, compared to human annotations, showcasing the efficacy of FMs in decomposing complex, multi-modal data.}
\end{itemize}


\section{Related Work}
\label{sec:Related_Work}
\textbf{Creating Natural Language Annotations for Robot Trajectories:}
Traditional approaches to generating language descriptions for robot trajectories rely heavily on human-created annotations \cite{shridhar2021cliport, shridhar2022perceiveractor, Lynch-RSS-21, NEURIPS2020_9909794d, pmlr-v164-jang22a, lynch2022interactive, brohan2023rt1}. 
In contrast, our work aims to automate this process. 
To this end, several prior works have developed frameworks that leverage automated techniques for generating task-level descriptions in robotic contexts. 
Notably, Ha et al. \cite{ha2023scaling} propose a framework that iteratively queries LLMs to break down high-level task instructions into sub-tasks with natural language (NL) descriptions. 
Our method could extend this approach by using a \hlgreen{Finite-State Machine} (FSM) controller to validate annotations, reducing the need for costly FM re-queries. 
Additionally, our semantic alignment metric, $\semanticscore$, enables a more detailed assessment of language quality compared to task performance metrics, which can overlook subtle nuances and noise in synthetic language data. 
Other FM-based methods automate trajectory labeling through object detection and heuristics \cite{blank2024scaling} or require extensive fine-tuning on human-labeled data \cite{xiao2022robotic}. 
By contrast, our approach relies on prompt-based in-context learning for FM reasoning, using human annotations solely to evaluate the $\SIMILARITY$ algorithm. 
Our method uniquely generates sub-task-level, temporally grounded annotations rather than descriptions for entire trajectories, enabling more precise alignment across trajectory steps.

\textbf{Semantic Preservation in Robotic Task Embeddings:}
Other work, such as Cao et al. \cite{Cao}, preserves the semantic structure of behavior trees (BTs) in robotic tasks by embedding the BT hierarchy and sequence into representations using \hlgreen{language embedding} models like USE \cite{cer2018USE} and GloVe \cite{GLOVE}. 
Unlike our approach, which focuses on assessing annotation quality rather than behavioral coordination, Cao’s method organizes descriptors within the BT structure. 
Their approach encodes sequential information through unitless exponential decay, lacking explicit temporal grounding unlike our method. However, we both employ cosine-based similarity on USE embeddings for validating semantics in robotic contexts.

Alternative approaches at the intersection of task planning and Explainable AI (XAI) \cite{XAI} use FMs to provide annotations for TAMP or generate summaries describing robot behavior \cite{dechant2022summarizing}. 
Some of these methods require highly detailed data, such as ROS2 log files \cite{LLMsinterprettingRobotBehaviors}, which limit their general applicability. 
Explainability is not the primary focus of our work, but is a potential future application of our method.

\textbf{Decomposing Trajectories}: \cite{raj2023languageconditioned} treat trajectory decomposition as semantic parsing and are similarly focused on detecting sub-tasks to train a detection algorithm with labeled video clips and language annotation pairs. 
Prior works in computer vision focus on similar concepts to alleviate manual annotation of video segments, \cite{joint_visual_temporal}, \cite{aakur2019perceptual}, \cite{sener2018unsupervised} which rely on training detection algorithms rather than prompting multi-modal FMs.

\textbf{Annotation Quality Assurance Metrics:} Previous works have primarily evaluated the quality of labeled data using binary success rates on downstream tasks. 
However, this binary approach overlooks important factors in data generation, such as coverage and diversity, which are critical when developing synthetic datasets. 
Assessing these factors can also highlight potential shortcomings, such as class imbalance and noise, e.g., temporally unaligned language instructions. 
Therefore, instead of relying on these binary metrics, we compare the task decomposition performance of our VLM and LLM models with that of human labelers.

\section{Problem Statement}
\label{sec:Problem_Statement}
\subsection{Preliminaries and Definitions}\label{ssec:preliminaries}
A trajectory, $\trajectory$, is a complete sequence of robot \textit{and} non-robot states, $\state \in \Stateset \subset \R^\statedim$, control inputs, $\control: \Stateset \mapsto \Controlset \subset \R^\controldim$, and 3 channel images, $\image \in \Imageset \subset \R^{H \times W \times 3}$, that describes the evolution of robot states and non-robot states over all time steps.
Robot states may encompass the robot's end-effector pose, whereas non-robot states may include the external objects pose.
We assume a trajectory is propagated forward by discrete-time dynamics with \hlyellow{step, $\step \in [0, \nsteps] \subset \N^{+}$}, as:
\begin{equation}
\begin{aligned}\label{eq:dynamics}
    \state_{\step+1} = f(\state_{\step}, \control_{\step}),
\end{aligned}
\end{equation}
where $f: \R^{\statedim} \times \R^{\controldim} \mapsto \R^{\statedim}$ serves as the dynamics for a robotic agent within a particular environment.
The $\step$ in $\state_\step$, $\control_\step$, and $\image_\step$ denotes the signal at step $\step$.
We denote the trajectory ``length'' as the total number of steps, $\nsteps$.
While $\image_{\step}$ is not directly in \cref{eq:dynamics}, $\image_{\step}$ captures a visual representation of the environment.
We consider the control policy to be a high-level TAMP controller, for example a FSM or a Behavior Tree (BT) \cite{BT_FSM_surveyOgren}, \cite{BT_robotcontrolsystems}.

\begin{definition}[\textbf{Trajectory Data}]\label{def:trajectorydata} A tuple of state, control, and an image recorded at every step, $\step$, that is generated by a trajectory is defined as:
    $$\trajectorydata = \{(\hlmath{\step}, \state_{\step}, \control_{\step}, \image_{\step})\}_{\step=0}^{\nsteps}.$$
\end{definition}
We denote $\textualdata \subset \trajectorydata$ as \textit{textual data}, which only contains $\{ (\hlmath{\step}, \state_\step, \control_\step)\}_{\step=0}^{\nsteps}$, and denote $\visualdata \subset \trajectorydata$ as \textit{visual data}, which only contains $\{\image_\step\}_{\step=0}^{\nsteps}$, so that $\textualdata \cup \visualdata = \trajectorydata$. 

The FM's prompt, $\prompt$, is the entire input to the FM, which incorporates $\trajectorydata$ and context, $\context$.
We will use the description for $\prompt$ and $\context$ explained in \cite{LLMs_fewshotlearners}.
The $\context$ section that provides the task to the FM is denoted herein as $\contextLLMtasking \subset \context$.
The $\context$ section that describes the data is denoted as $\contextdatadescription \subset \context$.
Importantly, a section of $\context$ can provide a One-Shot example, denoted herein as $\contextoneshot \subset \context$.
Zero examples, Zero-Shot, is denoted herein as $\contextzeroshot \subset \context$, and of course, $\contextoneshot$ and $\contextzeroshot$ are mutually exclusive in $\context$.

\begin{definition}[\textbf{Sub-task}]\label{def:subtasksegmemt}
An atomic and complete motion or process, bounded temporally and concisely described by a sentence, that collectively contributes to accomplishing a larger, complex goal is defined as: 
$$\subtask = (\startstep, \step_{\text{end}}, \subtaskdescription).$$
\end{definition}
A sub-task, $\subtask$, includes a start step, $\startstep$, end step, $\step_{\text{end}}$, and a natural language description, $\subtaskdescription$. 
Both $\startstep$ and $\step_{\text{end}}$ are the same step type as in \cref{def:trajectorydata}.
They delineate the \textit{temporal} bounds of the sub-task, establishing its commencement and completion within $\trajectory$. 
The sub-task description, $\subtaskdescription$, provides a concise, \textit{semantic} natural language description.

\begin{definition}[\textbf{Sub-task Decomposition}]\label{def:subtaskdecomposition}
A temporally ordered collection of sub-tasks, where each sub-task is indexed by $\idxsubtask \in [0, \nsubtasks] \subset \N^+$, defined as:
$$\Subtaskdecomposition = \{ (\startstep, \step_{\text{end}}, \subtaskdescription) \}_{\idxsubtask=0}^{\nsubtasks} = \{ \subtask_\idxsubtask \}_{\idxsubtask=0}^{\nsubtasks}.$$
\end{definition}
We denote $\nsteps_{\Subtaskdecomposition}$ and $\nsubtasks_{\Subtaskdecomposition}$ as the number of steps and number of sub-tasks, respectively, in $\Subtaskdecomposition$.
\hlyellow{Note that the map from $\trajectory$ to $\Subtaskdecomposition$ is not injective.}
\hlyellow{Practically, this means an $\Subtaskdecomposition$ for a given $\trajectory$ is determined by a designer who chooses language descriptors and the temporal partitions' granularity.}

\subsection{Robotics Manipulator Example}\label{ex:running_example}
Consider a robotic environment \texttt{Stack} that consists of a manipulator picking up Cube A, stacking Cube A on Cube B, then returning to a home position.
Each state at a step, $\state_{\step}$, consists of the robot's end-effector pose and the non-robot poses, which are each cubes' pose and the end-effector pose relative position to each cubes' pose.
The control signal at each step, $\control_{\step}$, is the desired manipulator joint pose.
The image at each step, $\image_{\step}$, is a $256 \times 256$ RGB image.
The task planner is an FSM, which carefully sequences sub-tasks together to achieve the goal of stacking Cube A on Cube B.

Let us assume the entire trajectory length is 62, so that $\nsteps_{\Subtaskdecomposition} = 62$, and the number of sub-tasks is 8, so that $\nsubtasks_{\Subtaskdecomposition} = 8$. 
After $\trajectorydata$ is recorded, an entire sub-task decomposition, $\Subtaskdecomposition$, can be written as:
\begin{align}\label{eq:subtaskdecomposition_example}
    \Subtaskdecomposition = \{ &(0, 10, \text{``Move to above Cube A''}), \\
                               &(11, 23, \text{``Move directly down to Cube A''}), \nonumber\\\nonumber
                               &\dots \\\nonumber
                               &(59, 62, \text{``Return home''}) \}. \nonumber
\end{align}

We will denote this $\Subtaskdecomposition$ as the ``ground-truth'' because the trajectory evolved from the FSM and therefore the resulting sub-task decomposition is imposed by the FSM; however, there can be multiple ground-truth $\Subtaskdecomposition$ for a given trajectory.
This entire $\Subtaskdecomposition$ is shown in the left column in \cref{tab:subtask_decomposition_examples}.

\subsection{Main Problems}\label{ssec:problem_statement}

\begin{problem}[\textbf{Steer an FM to Label a Trajectory}]
Design a prompt, $\prompt$, incorporating both $\context$ and $\trajectorydata$, for an FM to output a prediction sub-task decomposition, $\hat \Subtaskdecomposition$, that closely matches a ground-truth sub-task decomposition.
\end{problem}\label{prob:prompt} 

The FM is used ``as is'' with no retraining or fine-tuning to accurately form a sub-task decomposition prediction, $\hat{\Subtaskdecomposition}$, that aligns closely with ground-truth, $\Subtaskdecomposition$.
This method capitalizes on the FM's pre-trained knowledge base, allowing it to apply existing patterns and concepts to new, unseen tasks.
This approach enhances the efficiency and scalability of sub-task decomposition through the FM's applicability for automating nuanced, task-specific process definitions.

\begin{problem}[\textbf{Evaluate the FM Prediction}]\label{prob:evaluate}
Quantitatively assess the prediction quality, where the core of the evaluation focuses on temporal \textit{and} semantic alignment between ground-truth $\Subtaskdecomposition$ and FM prediction $\hat \Subtaskdecomposition$.
\end{problem}
 
Temporal alignment ensures $\hat \Subtaskdecomposition$ is the correct sequence and duration reflecting the dynamics of the task accurately. 
Meanwhile, semantic alignment evaluates the relevance and appropriateness of the language descriptions assigned to each sub-task, ensuring they accurately describe the underlying action or intention. 
Together, these measures comprehensively gauge the FM prediction effectiveness, ensuring they are accurate in timing and contextually meaningful.
\section{Methodology}
\label{sec:Approach}

\subsection{FMs for Auto-Decomposition of Robotic Sub-tasks}\label{ssec:prompt_engineer}
A main challenge lies in crafting a prompt, $\prompt$, with context, $\context$, that unlocks the balance between an FM's ``parametric'' knowledge \cite{rag}, analysis capability, and translation from $\prompt$ to $\hat \Subtaskdecomposition$.
Below is our method for building such a $\prompt$.

\begin{figure}[ht!]
    \centering
    \includegraphics[width=0.47\textwidth]{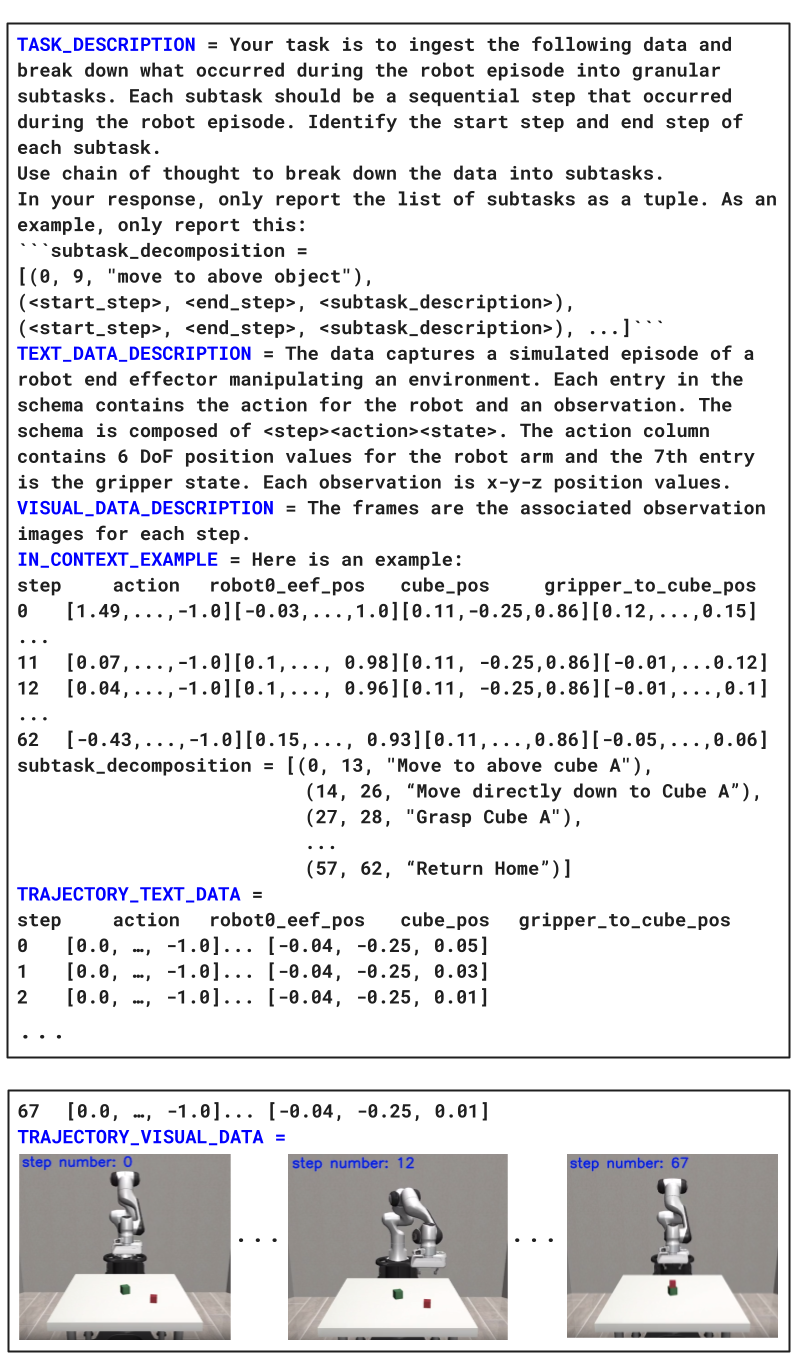}
    \caption{A Prompt, $\prompt$, with context, $\context$, and Trajectory Data, $\trajectorydata$, as described in \cref{ssec:prompt_engineer}. The key sections, in blue bold text, represent the different aspects of the $\prompt$, including the FM tasking, $\contextLLMtasking$; partial in-context example, $\contextoneshot$; textual data, $\textualdata$; and visual data, $\visualdata$.}
    \label{fig:prompt}
\end{figure}

The FM's task and response template is assigned through $\contextLLMtasking$, shown as \texttt{TASK\_DESCRIPTION} in \cref{fig:prompt}.
The trajectory data description, $\contextdatadescription$, briefly describes the expected $\trajectorydata$ schema, shown as \texttt{TEXT\_DATA\_DESCRIPTION} and \texttt{VIDEO\_DATA\_DESCRIPTION} in \cref{fig:prompt}.
The $\trajectorydata$ is placed in its entirety in the $\prompt$.

The core $\context$ piece shown in \cref{sec:Experiments} to enable high prediction quality is the ``One-Shot'' example, $\contextoneshot$, shown by \texttt{IN\_CONTEXT\_EXAMPLE} in \cref{fig:prompt}.
The $\contextoneshot$ is actually a \textit{snippet} of $\textualdata$ (from a hold-out $\trajectorydata$, $\Subtaskdecomposition$ pair) for each environment to provide only a partial example.
When \texttt{IN\_CONTEXT\_EXAMPLE} is omitted, we consider the $\context$ to include $\contextzeroshot$ rather than $\contextoneshot$.
\hlgreen{That is, $\contextzeroshot$ does not provide any in-context examples in the $\prompt$.}

To summarize, $\contextLLMtasking$ sets the FM's task and provides a response template, $\contextdatadescription$ briefly describes the $\trajectorydata$, $\contextoneshot$ provides a partial example, and  $\trajectorydata$ provides two input modalities through $\textualdata$ and $\visualdata$.

\subsection{Similarity Metrics Formulation}\label{ssec:metric}
We propose $\SIMILARITY$ in \cref{function:subtasksimilarity}, an algorithm that calculates metrics to address \cref{prob:evaluate} and compare the difference between $\Subtaskdecomposition$ and $\hat \Subtaskdecomposition$ through a temporal score, $\temporalscore$, and a semantic score, $\semanticscore$.
 $\SIMILARITY$ attributes include:
\begin{itemize}
    \item \hlyellow{$\temporalscore \in [0,1]$ and $\semanticscore \in [-1, 1]$, where 1 in both scoring metrics occurs only when sub-task decompositions are equal.} 
    \item Symmetry, such that the order of the arguments does not affect the result, i.e., $\SIMILARITY(\Subtaskdecomposition, \hat \Subtaskdecomposition) = \SIMILARITY(\hat\Subtaskdecomposition, \Subtaskdecomposition)$.
    \item Flexibility in semantic similarity by utilizing any pre-trained text encoding function.
    \item The ability to handle two inputs, $\Subtaskdecomposition$ and $\hat \Subtaskdecomposition$, that may contain a different number of sub-tasks, $\nsubtasks_{\Subtaskdecomposition} \neq \nsubtasks_{\hat{\Subtaskdecomposition}}$, and different number of steps, $\nsteps_{\Subtaskdecomposition} \neq \nsteps_{\hat{\Subtaskdecomposition}}$.
\end{itemize}
\textbf{Brief Algorithm Description}:
The algorithm compares each $\subtask_{\idxsubtask}$ in $\Subtaskdecomposition$ to each $\hat \subtask_{\idxsubtask}$ in $\hat \Subtaskdecomposition$.
Sub-tasks temporally intersect when the start and end indices of $\subtask_{\idxsubtask}$ are within the start and end indices of any $\hat \subtask_{\idxsubtask}$, implying an Intersection over Union (IOU) is greater than zero.
For all temporally intersecting sub-tasks the algorithm determines a temporal score, $\temporalscore$, based on IOU and adjusts this score by a weight, $\intervalweight_{\intersectingsubtaskidx}$, proportional to the overlap's significance to the trajectory length.
Similarly, the algorithm determines $\semanticscore$ adjusted by the same weight, $\intervalweight_{\intersectingsubtaskidx}$.

\begin{algorithm}
\caption{Sub-task Decomposition Similarity}
\label{function:subtasksimilarity}
\begin{algorithmic}[1]
\Function{SIMILARITY}{$\Subtaskdecomposition$, $\hat \Subtaskdecomposition$}
\State $\intersectingsubtaskidx$ $\gets$ 0 \Comment{Number of intersecting sub-tasks}
\For{each $\subtask_{\idxsubtask}$ in $\Subtaskdecomposition$}
    \For{each $\hat \subtask_{\idxsubtask}$ in $\hat \Subtaskdecomposition$}
        \If{\Call{iou}{$\subtask_{\idxsubtask}, \hat \subtask_{\idxsubtask}$} $> 0$} \label{algline:if_iou}
            \State $\intersectingsubtaskidx \gets \intersectingsubtaskidx + 1$
            \State $IOU_{\intersectingsubtaskidx} \gets \Call{iou}{\subtask_{\idxsubtask}, \hat \subtask_{\idxsubtask}}$ \label{algline:temporal_line}
            \State $CS_{\intersectingsubtaskidx} \gets \Call{cosine\_similarity}{\subtask_{\idxsubtask}, \hat \subtask_{\idxsubtask}}$ \label{algline:semantic_line}
            \State $\intervalweight_{\intersectingsubtaskidx} \gets \Call{interval\_weight}{\subtask_{\idxsubtask}, \hat \subtask_{\idxsubtask}}$ \label{algline:interval_weight_line}
        \EndIf
    \EndFor
\EndFor

\State $\temporalscore \gets \frac{\sum{IOU_\intersectingsubtaskidx} \times \intervalweight_\intersectingsubtaskidx}{\sum{\intervalweight_\intersectingsubtaskidx}}$ \label{algline:normalize_temporal} \Comment{Normalized temporal similarity}
\State $\semanticscore \gets \frac{\sum{CS_\intersectingsubtaskidx} \times \intervalweight_\intersectingsubtaskidx}{\sum{\intervalweight_\intersectingsubtaskidx}}$ \label{algline:normalize_semantic} \Comment{Normalized semantic similarity}
\State \textbf{return} $\temporalscore$, $\semanticscore$

\EndFunction
\end{algorithmic}
\end{algorithm}

\begin{table*}[ht!]
\centering
\begin{tabular}{@{}llll@{}}
\toprule
\makecell{Ground-truth $\Subtaskdecomposition$} 
& \makecell{$\hat \Subtaskdecomposition$ using $\textualdata$, $\contextoneshot$ \\ $\temporalscore = 0.87$, $\semanticscore = 0.98$ \\ cost = \$0.12 } 
& \makecell{$\hat \Subtaskdecomposition$ using  $\textualdata$, $\contextzeroshot$ \\ $\temporalscore = 0.74 $, $\semanticscore = 0.51$ \\ cost = \$0.10}
& \makecell{$\tilde \Subtaskdecomposition$  using $\visualdata$\\ $\temporalscore = 0.45$, $\semanticscore = 0.22$ \\
time spent = 133 seconds \\cost $\approx$ \$0.53  } \\
\midrule
\makecell[tl]{\{(0, 10, ``Move to above \\ \phantom{xxxxxxx} Cube A''), \\
(11, 23, ``Move directly down\\ \phantom{xxxxxxx} to Cube A''), \\
(24, 25, ``Grasp Cube A''), \\
(26, 39, ``Vertically pick \\ \phantom{xxxxxxx} up Cube A''), \\
(40, 48, ``Align Cube A \\ \phantom{xxxxxxx} with Cube B''), \\
(49, 54, ``Move Cube A \\ \phantom{xxxxxxx} vertically down \\ \phantom{xxxxxxx} to Cube B''), \\
(55, 58, ``Release Cube A \\ \phantom{xxxxxxx} onto Cube B''), \\
(59, 62, `Return Home'')\}} 
&\makecell[tl]{\{(0, 10, ``Move to above \\ \phantom{xxxxxxx} Cube A''),\\
(11, 23, ``Move directly \\\phantom{xxxxxxx} down to Cube A''), \\
(24, 25, ``Grasp Cube A''), \\
(26, 39, ``Vertically pick \\ \phantom{xxxxxxx} up Cube A''), \\
(40, 54, ``Align Cube A \\ \phantom{xxxxxxx}with Cube B''), \\
(55, 58, ``Release Cube A \\ \phantom{xxxxxxx} onto Cube B''), \\
(59, 62, ``Return Home')\}}
& \makecell[tl]{\{(0, 10, ``robot arm moving \\ \phantom{xxxxxxx} towards cube A''),\\
(11, 24, ``robot arm adjusting \\ \phantom{xxxxxxx} position above cube A''),\\
(25, 39, ``robot arm descending \\ \phantom{xxxxxxx} towards cube A''),\\
(40, 54, ``robot arm \\ \phantom{xxxxxxx} lifting cube A''),\\
(55, 62, ``robot arm \\ \phantom{xxxxxxx} moving cube A \\ \phantom{xxxxxxx} towards cube B'')\}} 
& \makecell[tl]{\{(0, 24, ``robot arm moves \\ \phantom{xxxxxxx} over red block'''),\\
(25, 27, ``robot arm grips \\ \phantom{xxxxxxx} red block''), \\
(28, 40 ``robot arm lifts \\ \phantom{xxxxxxx} red block''), \\
(41, 56, ``robot arm moves \\ \phantom{xxxxxxx} red block over \\ \phantom{xxxxxxx}  to the top of \\ \phantom{xxxxxxx} the green block''), \\
(57, 59, ``robot arm releases \\ \phantom{xxxxxxx} grip of red block'), \\
(60, 62, ``robot arm stows away'')\}} \\
\hline
\end{tabular}
\caption{\textbf{Ground-truth $\Subtaskdecomposition$ compared against two $\hat \Subtaskdecomposition$ and one $\tilde \Subtaskdecomposition$ for a single trajectory in the \texttt{Stack} environment: }
The second column shows a GPT-4V $\hat \Subtaskdecomposition$ using textual data, $\textualdata$, and an in-context example, $\contextoneshot$.
The third column shows a GPT-4V $\hat \Subtaskdecomposition$ using textual data, $\textualdata$, and excludes an in-context example, $\contextzeroshot$. 
The last column shows a human annotated sub-task decomposition, $\tilde \Subtaskdecomposition$.
The metrics, $\temporalscore$ and $\semanticscore$, shown at the top of each column are scores associated with $\SIMILARITY(\Subtaskdecomposition, \hat \Subtaskdecomposition)$ or $\SIMILARITY(\Subtaskdecomposition, \tilde \Subtaskdecomposition)$.
The key insight is $\hat \Subtaskdecomposition$ using $\contextoneshot$ is nearly identical to the ground-truth, while $\hat \Subtaskdecomposition$ using $\contextzeroshot$ has different temporal partitions and overall different vernacular.
Additionally, the two $\hat \Subtaskdecomposition$ are nearly 1/5 of the cost compared to $\tilde \Subtaskdecomposition$, and have higher $\temporalscore$ and $\semanticscore$ metrics.
}
\label{tab:subtask_decomposition_examples}
\end{table*}

\textbf{Example}: 
We will explain $\SIMILARITY$ with $\subtask_4$ in $\Subtaskdecomposition$ from \cref{eq:subtaskdecomposition_example} and $\hat \subtask_3$ in $\hat \Subtaskdecomposition$ that is produced by GPT-4V using $\textualdata$ and $\contextzeroshot$.
This $\hat \Subtaskdecomposition$ is fully shown in the third column of \cref{tab:subtask_decomposition_examples}.
The respective sub-tasks are:
\begin{align}
    \subtask_4 &= (40, 48,\text{``Align Cube A with Cube B''}) \nonumber \\
    \hat \subtask_3 &= (40, 54, \text{``robot arm lifting cube A''}).\nonumber
\end{align}

In both \cref{algline:if_iou} and \cref{algline:temporal_line}, the subroutine \texttt{IOU()} calculates the temporal alignment between $\subtask_4$ and $\hat \subtask_3$.
\begin{align}
    \texttt{IOU}(\subtask_4, \hat \subtask_3)=\texttt{IOU}((40,48),(40,54)) = 8/14 = 0.571\nonumber.
\end{align}

In \cref{algline:semantic_line}, the subroutine \texttt{COSINE\_SIMILARITY()} computes the semantic congruence between sub-task descriptions, $\subtaskdescription_4 = \text{``Align Cube A with Cube \hlgreen{B}''}$ and $\hat \subtaskdescription_3 = \text{``robot arm lifting cube A''}$.   
This assessment is achieved by calculating the cosine similarity between their encoded vector representations using the formula $\frac{v_1 \cdot v_2}{\left \| v_1 \right \|\left \| v_2 \right \|}$, where $v_1$ and $v_2$ are encoded vectors of $\subtaskdescription_4$ and $\hat \subtaskdescription_3$.
This yields a scalar between -1 (opposite) to 1 (most similar) with 0 indicating orthogonality.
We assume \texttt{COSINE\_SIMILARITY()} encodes embedded vectors of pre-trained semantics of sub-task descriptions written in natural language.
For our example two sub-tasks, the semantic relevancy between $\subtask_4$ and $\hat \subtask_3$ is:
\begin{equation}
\begin{aligned}
\texttt{COSINE\_SIMILARITY}(\subtask_4, \hat \subtask_3)\phantom{xxxxxxxxxxxxx} \nonumber\\
= \frac{\texttt{ENCODER}(\subtaskdescription_4) \cdot \texttt{ENCODER}(\hat\subtaskdescription_3)}{\left \| \texttt{ENCODER}(\nohat\subtaskdescription_4) \right \| \left \| \texttt{ENCODER}(\hat\subtaskdescription_3) \right \|} \nonumber
=0.504.
\end{aligned}
\end{equation}
\texttt{ENCODER()} represents any state-of-the-art pre-trained model capable of semantic encoding, such as USE \cite{cer2018USE}.
The results above use the USE encoder and we use the USE encoder in \cref{sec:Experiments}.
The choice of encoder is based on its efficacy in capturing textual semantics, essential for understanding and comparing sub-task descriptions \cite{chen2023difference}, \cite{radford2021learning}.
This approach ensures flexibility and adaptability in applying the most suitable semantic encoding technique for the task at hand.

In \cref{algline:interval_weight_line}, the subroutine \texttt{INTERVAL\_WEIGHT()} calculates the shortest combined window length of $\subtask_4$ and $\hat \subtask_3$, normalized by maximum trajectory lengths between $\nsteps_{\Subtaskdecomposition} $ and $\nsteps_{\hat{\Subtaskdecomposition}}$, which both happen to be 62 in this example.
The interval weight, $\intervalweight_\intersectingsubtaskidx$, between $\subtask_4$ and $\hat \subtask_3$ is:
\begin{align}
\texttt{INTERVAL\_WEIGHT}(\subtask_4, \hat \subtask_3)\phantom{xxxxxxxxxx}&  \nonumber\\
= \frac{\min(48,54) - \max(40,40) + 1}{\max(62,62)}& = 0.145. \nonumber
\end{align}
\hlgreen{The interval weight scales sub-task significance within the trajectory, preventing short overlaps from skewing similarity scores. Unlike IoU, it normalizes across trajectory lengths for consistency.}

After all sub-task intersections are calculated, \cref{algline:normalize_temporal} and \cref{algline:normalize_semantic} calculate $\temporalscore$ and $\semanticscore$ to be within their respective ranges, \hlgreen{i.e. $\temporalscore \in [0,1]$ and $\semanticscore \in [-1,1]$}.
\section{Experiments}
\label{sec:Experiments}
\begin{figure*}[ht!]
    \centering
    \includegraphics[width=\textwidth]{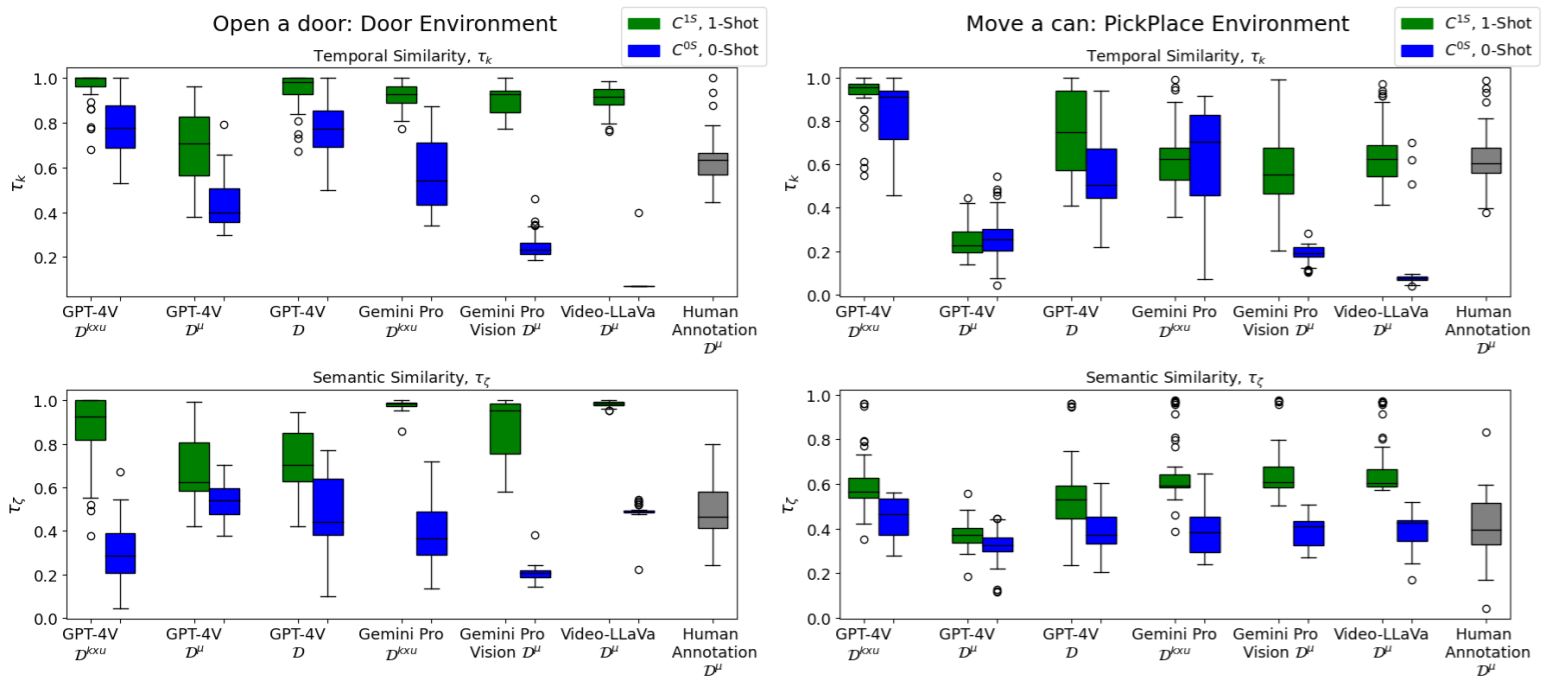}
    \caption{\textbf{Ablation study for two environments:} Temporal, $\temporalscore$, and semantic, $\semanticscore$, statistics for different FMs with varying input parameters including context $\context$, and data modalities (textual data, $\textualdata$, and visual data, $\visualdata$). 
    The key insight is an in-context example, $\contextoneshot$ (green), significantly boosts both $\temporalscore$ and $\semanticscore$, while visual data often decreases $\semanticscore$.
    Using our metrics across different FMs and $\prompt$, developers can choose the best FM configuration for their application. 
    In these examples GPT-4V with $\contextoneshot$ and $\textualdata$ performs the best and most consistent.}
    \label{fig:boxplots}
\end{figure*}
The goal of our experiments is to verify our approach on a large and diverse set of trajectory lengths, initial conditions, sub-task durations, and sub-task descriptions.
\hlyellow{Our experiments include 200 unique trajectories (4 environments, 50 trajectories per environment) with corresponding $\trajectorydata$ and a ground-truth $\Subtaskdecomposition$.}
All project code including experimental configurations, \cref{function:subtasksimilarity} implementation, FSM implementations, and results generation can be found in the github repository \href{https://github.com/jsalfity/task_decomposition}{github.com/jsalfity/task\_decomposition}.

All results are shown in \cref{tab:results_table}, which presents mean and standard deviation for both $\temporalscore$ and $\semanticscore$ across 50 trajectories (50 calculations of $\SIMILARITY(\Subtaskdecomposition, \hat \Subtaskdecomposition)$ for unique $\Subtaskdecomposition$) for 2400 different combinations of FMs, including GPT-4V, Gemini Pro, Gemini Pro Vision, Video-LLaVa; input modalities, $\textualdata$ and $\visualdata$; and contexts, $\contextoneshot$ and $\contextzeroshot$.

\subsection{Robotics' Trajectory Data Collection}
\hlgreen{\textbf{Robotics' Environments:}} We narrow the scope of our work to robot manipulators in the Robosuite \cite{robosuite2020} simulation environment.
\hlgreen{While the following experiments and results are limited to Robosuite, we note that any simulation environment or real-world experimental setup which can produce trajectory data, $\trajectorydata$, in the form defined in \cref{def:trajectorydata} can be similarly evaluated.}
The 4 different environments used in this work are seen in \cref{tab:results_table} including their approximate $\nsteps_{\Subtaskdecomposition}$ and $\nsubtasks_{\Subtaskdecomposition}$.
The environments range from short, single objective tasks to longer, complicated multi-objective tasks.

\hlgreen{\textbf{Finite-State Machine: }}\hlyellow{All environments are controlled using FSM-based control, seen in the supporting code, to control robot's end-effector position based on pre-defined sub-tasks.
The transition between sub-tasks is determined by spatial position conditions such as when a gripper is directly above an object, when a grasp is achieved, etc.
Each sub-task is labeled directly by the FSM.
As an example, the \texttt{Lift} environment's sequential sub-task descriptors are: ``Move to cube'', ``Grasp Cube'', ``Lift Cube'', and moves to the next sub-task when the previous sub-task is complete.
The FSM approach enables tracking specific sub-tasks \textit{and} the associated start/stop steps.
In each sub-task, a position feedback controller commands the end-effector to move to the desired x-y-z position using Robosuite's \texttt{OSC\_POSE} controller.}

The control vector is $\control \in \R^7$ and includes the desired end-effector pose in $\R^6$ and a boolean value that specifies the gripper state (open or closed).
The state vector, $\state$, consists of the global \textit{robot} state x-y-z end-effector positions in $\R^3$ and global \textit{non-robot} object states, where each object is represented by x-y-z positions in $\R^3$.
Examples of non-robot objects are cubes, door handles, cans, etc.
The $\visualdata$ consists of 256 $\times$ 256 RGB images, encoded into Base64 format, with the step number annotated at the top of each image, as shown in the bottom of the $\prompt$ in \cref{fig:prompt}.

Environmental objects are randomly positioned to generate unique trajectories.
The general manipulation task for each environment remain similar but each sub-task contains different start and end steps and different $\nsteps_{\Subtaskdecomposition}$.
The $\nsubtasks_{\Subtaskdecomposition}$ and approximate $\nsteps_{\Subtaskdecomposition}$ are written below each environment in \cref{tab:results_table}.

\hlgreen{\textbf{Ablation:}} Although all this information is available to us through the simulated environment, we conduct an ablation study in our experiments, \cref{fig:boxplots}, to vary the amount of information provided to each FM for sub-task decomposition. 
Notably, $\visualdata$, which represents scenarios where only RGB camera data is available for labeling, best reflects real-world use cases. 

\hlyellow{\textbf{Human Annotations:}
$\SIMILARITY(\Subtaskdecomposition, \tilde \Subtaskdecomposition)$ statistics with ground-truth, $\Subtaskdecomposition$, and human annotated sub-task decompositions, $\tilde \Subtaskdecomposition$, provide a baseline comparison.
Human annotators in our experiments constitute roboticists from the authors' academic group and are not biased by the authors.}
\hlgreen{Human annotations are collected by presenting annotators with only visual data, $\visualdata$, i.e. the video, and similar instructions as shown in \cref{fig:prompt}.}
\hlyellow{A partial in-context example is shown for each video, which guides the $\tilde \Subtaskdecomposition$ granularity and vernacular.
As each human annotation is timed, we learned constructing a $\tilde \Subtaskdecomposition$ takes 2-3 minutes, which is nearly 5x the cost of an FM-based annotation.
An example $\tilde \Subtaskdecomposition$ is shown at the right-most column of \cref{tab:subtask_decomposition_examples} and statistics are in \cref{tab:results_table} and \cref{fig:boxplots}.}

\subsection{Querying an FM}
Four FM models are used: GPT-4V \cite{openai2023gpt4vision}, Gemini Pro, Gemini Pro Vision \cite{geminiteam2023gemini}, and Video-LLaVA \cite{lin2023videollava}.
The configuration settings for each model can be found in \cref{sec:Appendix} and the supporting code.
These models offer a diversity of input modality (textual and visual) and different parametric tuning accessibility and customization, e.g., Video-LLaVA is open source, whereas GPT-4V is only accessible via API.

For each FM query, the $\prompt$ is constructed as shown in \cref{fig:prompt}, with variations only in $\contextoneshot$ for each environment and $\trajectorydata$ for each trajectory. The same $\contextoneshot$ is used across environments and is based on a hold-out pair of $\trajectorydata$ and $\Subtaskdecomposition$, capturing key transitions between sequential sub-task decompositions.

\hlgreen{\textbf{Statistics:}} \hlyellow{ \cref{tab:results_table} shows statistics over FM responses with $\hat \Subtaskdecomposition$ that are valid according to \cref{def:subtaskdecomposition}.}
The error rates for each batch of 50 FM calls, due to 50 trajectories per environment, is reported in \cref{tab:LLM_error_rates}, \hlgreen{which is the corresponding sample size for the mean and standard deviation of each $\temporalscore, \semanticscore$ pair in \cref{tab:results_table}.}
Examples of invalid FM responses include when $\startstep$ comes after $\step_{\text{end}}$, $\startstep$ in any $\hat \subtask_{\idxsubtask}$ is greater than $\startstep$ in any $\hat \subtask_{\idxsubtask+1}$, or the FM response contains no $\hat \Subtaskdecomposition$ at all.
\hlgreen{Gemini Pro can only take in textual inputs}; Gemini Pro Vision and Video-LLaVa could not accept $\trajectorydata$ due to their respective $\prompt$ token limit and therefore only have $\visualdata$ statistics.

\setlength{\tabcolsep}{5pt}

\begin{table*}[ht]
\centering
\begin{tabular}{@{}lcccccccc@{}}
\toprule
& \multicolumn{2}{c}{Open a door: \texttt{Door}}
& \multicolumn{2}{c}{Lift a cube: \texttt{Lift}} 
& \multicolumn{2}{c}{Move a can: \texttt{PickPlace}}
& \multicolumn{2}{c}{Stack two cubes: \texttt{Stack}} \\

& \multicolumn{2}{c}{$\nsteps_{\Subtaskdecomposition} \approx 80$, $\nsubtasks_{\Subtaskdecomposition} = 5$}
& \multicolumn{2}{c}{$\nsteps_{\Subtaskdecomposition} \approx 40$, $\nsubtasks_{\Subtaskdecomposition} = 3$}
& \multicolumn{2}{c}{$\nsteps_{\Subtaskdecomposition} \approx 80$, $\nsubtasks_{\Subtaskdecomposition} = 7$}
& \multicolumn{2}{c}{$\nsteps_{\Subtaskdecomposition} \approx 120$, $\nsubtasks_{\Subtaskdecomposition} = 5$}\\
\cmidrule(r){2-3} \cmidrule(lr){4-5} \cmidrule(lr){6-7} \cmidrule(l){8-9}
\textbf{FM (Context, Data)} 
& \makecell{\textbf{$\temporalscore$}} & \makecell{\textbf{$\semanticscore$}}
& \makecell{\textbf{$\temporalscore$}} & \makecell{\textbf{$\semanticscore$}} 
& \makecell{\textbf{$\temporalscore$}} & \makecell{\textbf{$\semanticscore$}} 
& \makecell{\textbf{$\temporalscore$}} & \makecell{\textbf{$\semanticscore$}} \\
\midrule
Human Annotations $(\visualdata)$ & $0.64 \pm 0.11$ &$0.48 \pm 0.12$ &$0.81 \pm 0.12$ &$0.45 \pm 0.14$ &$0.63 \pm 0.14$ &$0.41 \pm 0.14$ &$0.66 \pm 0.16$ &$0.38 \pm 0.13$\\
\midrule
GPT-4V $(\contextoneshot, \textualdata)$ &$\textbf{0.97} \pm \textbf{0.07}$ &$0.87 \pm 0.16$ &$0.80 \pm 0.15$ &$\textbf{0.83} \pm \textbf{0.10}$ &$\textbf{0.93} \pm \textbf{0.10}$ &$0.60 \pm 0.12$ &$\textbf{0.74} \pm \textbf{0.12}$ &$\textbf{0.90} \pm \textbf{0.08}$ \\
GPT-4V $(\contextzeroshot, \textualdata)$ &$0.78 \pm 0.13$ &$0.30 \pm 0.13$ &$0.81 \pm 0.16$ &$0.61 \pm 0.14$ &$0.83 \pm 0.14$ &$0.45 \pm 0.09$ &$0.71 \pm 0.11$ &$0.32 \pm 0.19$ \\
\midrule
GPT-4V $(\contextoneshot, \visualdata)$ &$0.70 \pm 0.17$ &$0.69 \pm 0.18$ &$0.32 \pm 0.14$ &$0.48 \pm 0.18$ &$0.25 \pm 0.07$ &$0.37 \pm 0.06$ &$0.53 \pm 0.15$ &$0.88 \pm 0.06$ \\
GPT-4V $(\contextzeroshot, \visualdata)$ &$0.44 \pm 0.10$ &$0.53 \pm 0.07$ &$0.32 \pm 0.11$ &$0.20 \pm 0.04$ &$0.25 \pm 0.11$ &$0.32 \pm 0.07$ &$0.40 \pm 0.08$ &$0.51 \pm 0.10$\\

\midrule
GPT-4V $(\contextoneshot, \trajectorydata)$ &$0.95 \pm 0.08$ &$0.71 \pm 0.14$ &$0.78 \pm 0.14$ &$0.76 \pm 0.05$ &$0.75 \pm 0.20$ &$0.54 \pm 0.16$ &$0.70 \pm 0.12$ &$0.75 \pm 0.09$\\
GPT-4V $(\contextzeroshot, \trajectorydata)$ &$0.77 \pm 0.13$ &$0.47 \pm 0.17$ &$0.73 \pm 0.12$ &$0.57 \pm 0.15$ &$0.57 \pm 0.19$ &$0.39 \pm 0.09$ &$0.71 \pm 0.13$ &$0.62 \pm 0.10$\\
\midrule

Gemini Pro $(\contextoneshot, \textualdata)$ &$0.92 \pm 0.06$ &$\textbf{0.98} \pm \textbf{0.02}$ &$\textbf{0.84} \pm \textbf{0.15}$ &$0.52 \pm 0.21$ &$0.62 \pm 0.13$ &$0.65 \pm 0.14$ &$0.53 \pm 0.15$ &$0.88 \pm 0.05$\\
Gemini Pro $(\contextzeroshot, \textualdata)$ &$0.57 \pm 0.15$ &$0.40 \pm 0.14$ &$0.64 \pm 0.15$ &$0.38 \pm 0.09$ &$0.64 \pm 0.21$ &$0.40 \pm 0.12$ &$0.48 \pm 0.11$ &$0.31 \pm 0.10$\\
\midrule
Gemini Pro V $(\contextoneshot, \visualdata)$ & $0.91 \pm 0.07$ &$0.84 \pm 0.15$ &$0.61 \pm 0.18$ &$0.74 \pm 0.24$ &$0.57 \pm 0.19$ &$\textbf{0.67} \pm \textbf{0.14}$ &$0.54 \pm 0.17$ &$0.89 \pm 0.06$  \\
Gemini Pro V $(\contextzeroshot, \visualdata)$ & $0.25 \pm 0.05$ &$0.21 \pm 0.03$ &$0.17 \pm 0.04$ &$0.28 \pm 0.03$ &$0.19 \pm 0.04$ &$0.38 \pm 0.06$ &$0.31 \pm 0.04$ &$0.24 \pm 0.04$ \\
\midrule
Video-LLaVA $(\contextoneshot, \visualdata)$ &$0.91 \pm 0.06$ &$\textbf{0.98} \pm \textbf{0.01}$ &$0.58 \pm 0.11$ &$0.65 \pm 0.02$ &$0.64 \pm 0.13$ &$\textbf{0.67} \pm \textbf{0.14}$ &$0.32 \pm 0.17$ &$0.54 \pm 0.01$ \\
Video-LLaVA $(\contextzeroshot, \visualdata)$ & $0.08 \pm 0.05$ &$0.49 \pm 0.04$ &$0.34 \pm 0.10$ &$0.34 \pm 0.02$ &$0.11 \pm 0.13$ &$0.39 \pm 0.07$ &$0.18 \pm 0.08$ &$0.28 \pm 0.04$\\
\bottomrule
\end{tabular}
\caption{
\textbf{Ablation study across all environments:} Comparison of temporal similarity, $\temporalscore$, and semantic similarity, $\semanticscore$, for different robotics manipulation environments ablated across prompts, $\prompt$, with different inputs modalities (textual data, $\textualdata$ and visual data, $\visualdata$) and context, $\context$, (in-context example, $\contextoneshot$, and no in-context example, $\contextzeroshot$).
Statistics (mean and standard deviation) are reported over approximately 50 predictions for each environment, FM, input modality, and context combinations. 
The key insight is the significant increase in temporal and semantic similarities for $\hat \Subtaskdecomposition$ with $\prompt$ that include $\textualdata$ and $\contextoneshot$. Statistics are in bold highlight exemplary high values for each metric and environment. }
\label{tab:results_table}
\end{table*}

\subsection{Results}
Our experimental results, seen completely in \Cref{tab:results_table}, show that FM-generated sub-task decompositions achieve high temporal, $\temporalscore$, and semantic $\semanticscore$, similarity scores when guided with an in-context example, $\contextoneshot$. 
This is because $\contextoneshot$ directly steers the FM to attend to important transitional steps and to synthesize descriptors using terminology similar to the example.

For example, GPT-4V with textual input, $\textualdata$, significantly outperforms both its zero-shot counterpart $\contextzeroshot$ and human annotations.
\hlgreen{The ablation study further reveals that including an in-context example not only improves alignment metrics but also curbs overly fine-grained predictions, while using visual data alone $\visualdata$ tends to reduce semantic fidelity.}
The very low scores are consistent for $\hat \Subtaskdecomposition$ that contain $\nsubtasks_{\hat \Subtaskdecomposition} \gg \nsubtasks_{\Subtaskdecomposition}$, indicating too fine of a granularity.
For example, Video-LLaVa produced $\nsubtasks_{\hat \Subtaskdecomposition} \approx \nsteps_{\Subtaskdecomposition}$, implying a sub-task at every $\step$, which is not realistic.

\hlgreen{In practice, our FM-based annotation framework enables dataset curators to validate only a small set of FM-generated annotations against hand-annotated trajectories, dramatically reducing annotation time and cost.
This approach facilitates the scalable production of high-quality, language-supervised datasets.}
\begin{table}[ht!]
\centering
\begin{tabular}{@{}lcccc@{}}
\toprule
\textbf{LLM (Context, Data)} 
& \makecell{\texttt{Door}} 
& \makecell{\texttt{Lift}} 
& \makecell{\texttt{PickPlace}} 
& \makecell{\texttt{Stack}} \\
\midrule
GPT-4V $(\contextoneshot, \textualdata)$ & 50 & 50 & 50 & 49\\
GPT-4V $(\contextzeroshot, \textualdata)$ & 50 & 50 & 50 & 50\\
\midrule
GPT-4V $(\contextoneshot, \visualdata)$ & 50 & 49 & 50 & 50\\
GPT-4V $(\contextzeroshot, \visualdata)$ & 50 & 50 & 50 & 50\\
\midrule

GPT-4V $(\contextoneshot, \trajectorydata)$ & 50 & 50 & 50 & 49\\
GPT-4V $(\contextzeroshot, \trajectorydata)$ & 49 & 50 & 50 & 50\\
\midrule

Gemini Pro $(\contextoneshot, \textualdata)$ & 42 & 48 & 50 & 50\\
Gemini Pro $(\contextzeroshot, \textualdata)$ & 47 & 43 & 47 & 49\\
\midrule
Gemini Pro V $(\contextoneshot, \visualdata)$ & 23 & 50 & 29 & 19\\
Gemini Pro V $(\contextzeroshot, \visualdata)$ & 47 & 34 & 44 & 38\\
\midrule
Video-LLaVA $(\contextoneshot, \visualdata)$ & 50 & 50 & 50 & 12\\
Video-LLaVA $(\contextzeroshot, \visualdata)$ & 50 & 50 & 50 & 50\\
\bottomrule
\end{tabular}
\caption{
\hlyellow{The number of correctly formatted $\hat \Subtaskdecomposition$, according to \cref{def:subtaskdecomposition}, for each 50 trajectory batch.
The number in each cell is the corresponding sample size for mean and standard deviation statistics of each $\temporalscore$, $\semanticscore$ pair in \cref{tab:results_table}.}
}\label{tab:LLM_error_rates}
\end{table}

\section{Conclusion}
We present a method to send multi-modal input data including step-by-step state, action and corresponding images to FMs including LLMs and VLMs.
We steer an FM for use as an analytics engine to digest raw trajectory data in the prompt and query the FM to decompose the data into distinct sub-tasks.
As of this writing, the cost of offloading and requesting data is \$0.03 to \$0.15 depending on the trajectory length.
For a human annotator to be more cost-effective, they would need to annotate a trajectory in about 15-30 seconds, which is not consistent with our collected human annotation data of 2-3 minutes per trajectory.
We see this analysis and capability of using FMs for robotic's data labeling as cost-effective and data efficient for the broader research problem of robot learning at scale.

The results are shown with a ground-truth $\Subtaskdecomposition$ crafted by a customized FSM.
There may be other $\Subtaskdecomposition$ with different granularity of sub-tasks, resulting in a different number of sub-tasks, $\nsubtasks_{\Subtaskdecomposition}$, and alternative descriptors.
Our results are valid because the proposed annotation pipeline and evaluation method will be set up by developers who want FM predictions, $\hat \Subtaskdecomposition$, and similar to $\Subtaskdecomposition$ of their own.

The IOU calculation is a strict temporal calculation, resulting in strict weightings.
If this causes challenges for future developers, an alternative ``soft'' IOU may be utilized where an intersection may be greater than zero for temporal ranges that are near each other, but do not necessarily intersect.
\hlyellow{In our algorithm, shorter sub-tasks are assigned smaller weights towards $\temporalscore$ and $\semanticscore$.
In practice, a short sub-task may be more important to a developer than longer sub-tasks.
Users can manually increase the relative weighting of important sub-tasks to give more contribution towards the final scores.}
While we use the USE encoder to calculate $\semanticscore$, our framework is designed to be modular, allowing future use of \textit{any} pre-trained encoder.

The presented robotics manipulator environments have structured and repeatable $\Subtaskdecomposition$, which allow us to present quantitative results.
The presented simulated environments offer perfect robot and non-robot pose information, which may not be assumed in the real-world.
We believe our experiments using only $\visualdata$ input, which does not contain pose information, closely represents real-world conditions.

Experimentation can be extended in several manners: additional visual data inputs per trajectory through multiple camera perspectives, physical robots with noisy positional data, non-manipulator environments, etc.
However, we believe the presented experiments and quantitative results are sufficient to showcase our proposed method.
\bibliographystyle{IEEEtran}
\bibliography{references}
\section{Appendix}
\label{sec:Appendix}
GPT-4V run with \texttt{gpt-4-vision-preview} in \texttt{ChatCompletion()} using OpenAI's API. 
Gemini Pro run with \texttt{gemini-pro} in \texttt{generate\_content()} using Google's Generative AI API.
Gemini Pro Vision run \texttt{gemini-pro-vision} in \texttt{generate\_content()} with Google's VertexAI API.
Video-LLaVa (language bind 7B) \cite{lin2023videollava} run locally on NVIDIA A100 80GB GPU.

\end{document}